\documentclass{article} 
\usepackage{collas2022_conference,times}
\usepackage{graphicx}
\usepackage{xspace}
\newcommand{\ourapproach}{\emph{CLAMP}\xspace}
\usepackage{subfig}

\usepackage{amsmath,amsfonts,bm}

\def\eqref#1{equation~\ref{#1}}

\def\1{\bm{1}}

\DeclareMathAlphabet{\mathsfit}{\encodingdefault}{\sfdefault}{m}{sl}
\SetMathAlphabet{\mathsfit}{bold}{\encodingdefault}{\sfdefault}{bx}{n}

\usepackage{hyperref}
\hypersetup{
    colorlinks=true,
    linkcolor=red,
    filecolor=magenta,      
    urlcolor=blue,
    citecolor=purple,
    pdftitle={Overleaf Example},
    pdfpagemode=FullScreen,
    }

\title{Latent Properties of Lifelong Learning Systems}

\author{Corban Rivera, Chace Ashcraft, Alexander New, James Schmidt, Gautam Vallabha \\
Intelligent Systems Center\\
Johns Hopkins Applied Physics Lab\\
\texttt{\{corban.rivera\}@jhuapl.edu} \\
}

\collasfinalcopy 

\begin{document}

\maketitle

\begin{abstract}
Creating artificial intelligence (AI) systems capable of demonstrating lifelong learning is a fundamental challenge, and many approaches and metrics have been proposed to analyze algorithmic properties. However, for existing lifelong learning metrics, algorithmic contributions are confounded by task and scenario structure. To mitigate this issue, we introduce an algorithm-agnostic explainable surrogate-modeling approach to estimate latent properties of lifelong learning algorithms. We validate the approach for estimating these properties via experiments on synthetic data.  To validate the structure of the surrogate model, we analyze real performance data from a collection of popular lifelong learning approaches and baselines adapted for lifelong classification and lifelong reinforcement learning.  
\end{abstract}

\section{Introduction}

Inspired by the way that humans acquire new skills sequentially and improve over time, lifelong or continual learning ~(\cite{Chen2018Book,Silver2013Lifelong}) describes the goal of enabling AI systems to learn tasks sequentially over time while improving performance on both previous and future tasks.  Lifelong learning has received much attention in the AI community, and many algorithms have been proposed for both supervised~(\cite{delange2020continual}) and reinforcement learning~(\cite{khetarpal2020rl_perspective}). We include additional review of lifelong learning approaches in Appendix \ref{sec:lifelong}. A key challenge, apart from the learning algorithm itself, is the assessment of the learner: how well it works, where it fails, and what factors influence its success or failure. Many existing approaches focus on defining continual learning metrics, such as \emph{forgetting}, \emph{backward transfer}, and \emph{forward transfer} ~(\cite{diaz2018dontforget,powers2021cora}); recently a suite of metrics for continual learning has been proposed~\cite{New2022metrics}. However, a metric such as forward transfer confounds three factors: the relationship between the tasks to be learned, the particular sequencing of tasks (curriculum), and the capabilities of the learner~(\cite{farquhar2019evaluations}). Use of systematic benchmarks (e.g., ~\cite{powers2021cora,lomonaco2017core50}) can address part of this concern by fixing the tasks and curricula, but does not address how the learner might do on other (non-benchmark) tasks.


In this work, we present a novel approach we refer to as Continual Learning Analysis via a Model of Performance, or \ourapproach, to estimate the separate contributions of task structure and algorithm capabilities from performance data. We refer to these contributions as \textit{latent properties} by analogy to latent variable modeling, where the goal is to identify a lower-dimensional space that effectively captures the probability distribution of high-dimensional data. In our case, we treat the performance curves (the time series of the learner's performance over the curriculum) as the high-dimensional data, and consider how to identify underlying properties of the learner and tasks that can best explain the observed performance curves. 
Our strategy is to (a) define a generative surrogate performance model of lifelong learning that has a small set of explainable parameters, and (b) estimate these parameters of the surrogate model from a set of performance curves. The explainable parameters specify algorithm properties (transfer efficiency, skill retention, expertise translation) and task properties (the ``similarity'' between the tasks in the curriculum). Our approach works equally well with classification learners and reinforcement learners, and relies solely on the performance curves (i.e., it does not need access to the implementation of the learner or the original task environment, similar to other post-hoc ML model assessment approaches~(\cite{Martin2021no_access})).  Additional review of explainable surrogate models literature is included in Appendix \ref{sec:explainable}.


\section{Approach}
Here we introduce Continual Learning Analysis via a Model of Performance (\ourapproach) and define the latent properties being estimated by the approach.  This work represents what we believe is the first explainable surrogate performance model for lifelong learning system analysis. We further believe this is the first model-agnostic attempt to quantitatively characterize both algorithm capabilities and inter-task relationships using only performance data available from multiple continual learning algorithms and multiple tasks. 

\subsection{Definitions}\label{sec:definitions}

A lifelong learning system (called the agent) experiences $n$ tasks ${t_1,t_2,\ldots,t_n}$ from an set $\mathcal{T}$. Formally, task experiences are tuples $(x,y) \in \mathcal{X}\times\mathcal{Y}$ for an input set $\mathcal{X}$ and output set $\mathcal{Y}$.  
We assume task experiences are ordered and known. Agents experience a sequence of $m$ experience tuples $(x_1,y_1),\ldots,(x_m,y_m)$ drawn from a paired sequence of tasks $\mathcal{C} = [c_1,c_2,\ldots,c_m]$, called a curriculum, where each $c_j \in \mathcal{T}$.  While single-task learning problems suppose that $(x_j,y_j)\sim\mathbb{P}_{\mathcal{X}\times \mathcal{Y}}$, sampling with curriculum supposes, by contrast, that $(x_j,y_j)\sim\mathbb{P}^{c_j}_{\mathcal{X}\times\mathcal{Y}}$ for $c_j\in \mathcal{T}$.\footnote{Formally, let $S:=\{(x_1,y_1),\ldots,(x_m,y_m)\}\subset \mathcal{X}\times\mathcal{Y}$ be a labeled data set, and $S^j=\{(x_i,y_i)\in S:\, c_i =c_j\}$; then we suppose $(x,y)\sim\mathbb{P}^{c_j}_{\mathcal{X}\times \mathcal{Y}}$ for $(x,y) \in S^j$.} As noted in Appendix~\ref{sec:categories}, task incremental learning includes task $c_j$ along with labeled data $(x_j,y_j)$ for the learning algorithm, while domain incremental learning does not. 

We consider two categories of properties. Intrinsic-task properties characterize a task to be encountered and are assumed to be invariant to changes in the lifelong learning algorithm encountering that task. Likewise, intrinsic-algorithm properties characterize properties of a given lifelong learning algorithm and are assumed to be invariant to what tasks that algorithm encounters. These strong assumptions enable us to ensure that \ourapproach\,\,is explainable.

The \textbf{task transfer matrix} $\textbf{A}\in \mathbb{R}^{n\times n}$ (recall that $|\mathcal{T}| = n$) with entries $\mathbf{A}_{i,j} \in [-1, 1]$ indicating how gaining experience from task $i$ affects performance on task $j$ independent of the algorithm selected. We assume a linear model of task transfer.

The \textbf{task difficulty score} $d$ in range $[0,\infty)$ as a task associated parameter that represents the intrinsic difficulty in increasing performance on a task given experience.

The \textbf{transfer efficiency score} $\gamma$ is an intrinsic property of lifelong learning algorithms in the range $[0,\infty)$ that indicates how efficiently the algorithm translates experience from one task to experience on another.

The \textbf{experience retention score} $h$ as a property of lifelong learning algorithms that conveys how well the algorithm retains prior experiences.  Values are in the range $[0,1]$ with $0$ indicating complete forgetting of prior task knowledge and $1$ indicating complete retention of prior task performance.

The \textbf{expertise translation score} $\lambda$ to be an algorithm property reflecting the ability of the algorithm to translate performance from one task into experience on another task, for any fixed task transfer matrix.  Values are in the range $[0,\infty)$.

\subsection{Lifelong Surrogate Model Formulation}

We assume a functional form for experience accumulation.  For an algorithm $a$ and curriculum $\mathcal{C}$, accumulated experience for position $i$ in curriculum $\mathcal{C}$, we define experience $\mathcal{E}^a_j(c_l)$ of task $t_j$ at all points $l$ in the curriculum as follows.  For the base case, we assume for all $t_j \in \mathcal{T}$ that the initial experience for all tasks is zero.
$\mathcal{E}^a_j(c_0) = 0 $
For curriculum steps $j>0$ we define experience accumulation by the following expression:
\begin{align*}
    \mathcal{E}^a_j(c_l) = \mathcal{E}(\mathcal{C},\textbf{A},\gamma,h,\lambda ) 
    = \mathcal{E}^a_j(c_{l-k})h_a + \textbf{A}_{i,j}(\gamma_a+\mathcal{P}_i(c_{l-1})\lambda_a),
\end{align*}%
where $\mathbf{A},\gamma, h,\lambda$ are task and algorithm properties defined in Section~\ref{sec:definitions}.  Experience in task $t_j$ is mapped to performance $\mathcal{P}(c_j)$ in task $t_j$ through a sigmoid function $S(x)=(1+e^{-x})^{-1}$, which is then shifted and scaled so that the inflection passes through zero.  If curriculum step $c_l$ experiences task $t_j$, then its performance $\mathcal{P}_i(c_l)$ is given by:
\begin{align*}
    \mathcal{P}_j(c_l) = S(\mathcal{E}_j(c_l)/d_i)
    = 2/(1+e^{-((\mathcal{E}_j(c_l)/d_i))})-1
\end{align*}

\subsection{Estimating parameters from data}\label{sec:estimation}

Suppose we have a curriculum $\mathcal{C}$ containing $m$ encounters of $n$ different tasks, and, for a set of algorithms $a = 1,\hdots,p$, a time-series of performance values $P^a \in \mathbb{R}^{n\times m}$, where the entries $P^a_{jl}$ are the performance of the $j$th task at the $l$th curriculum entry.  We desire to estimate the task transfer matrix $\mathbf{A}$ and task difficulty score $d$, as well as, for each algorithm $a$, the transfer efficiency $\gamma_a$, experience retention $h_a$, and expertise translation $\lambda_a$. We can group these parameters together as $\Theta = \{\mathbf{A},d\}\cup\bigcup_a \{\gamma_a, h_a, \lambda_a\}$. Then \ourapproach\,\,uses $\mathbf{A}, d, \gamma_a, h_a$, and $\lambda_a$ to predict a time-series of performance values $\hat{P}^a \in \mathbb{R}^{n\times m}$, where $\hat{P}^a_{jl}$ is the predicted performance of the $j$th task at the $l$th curriculum entry. Our parameters $\Theta$ can be estimated by solving the following minimization problem:

\begin{equation*}
\begin{aligned}
\min_{\Theta}\quad&\sum_a||\hat{P}^a - P^a||_F^2\\
\textrm{s.t.}\quad&-1 \leq \mathbf{A}_{ij} \leq 1\\
&d,\gamma_a,\lambda_a \geq 0\\
&0 \leq h_a \leq 1,
\end{aligned}
\end{equation*}

This minimization problem is differentiable in its parameters and has linear constraints, so it may be solved with techniques like projected subgradient descent. In practice, we implement our approach in PyTorch~\cite{pytorch} and use 1,000 steps of gradient descent with an Adam optimizer \cite{Kingma2014adam} with default parameters and learning rate. After each step, the parameters are projected back onto their feasible set.

\section{Results}
We designed experiments to (i) validate the approach used to estimate latent properties of lifelong learning, (ii) validate the functional form of the surrogate performance model with data coming from multiple baseline algorithms and multiple baseline datasets; including MNIST, CIFAR100 and Atari.  Due to space constraints, several sections are included in the appendices including: quantitative validation of \ourapproach based on synthetic lifelong learning performance data in Appendix  \ref{sec:synthetic}, analysis of domain adaptation with MNIST in Appendix \ref{sec:mnist}, and task adaptation with CIFAR100 in Appendix \ref{sec:cifar100}.

\label{sec:atari}
To illustrate the adaptability of \ourapproach, we analysed the performance of various lifelong learning reinforcement learning approaches from an experiment consisting of a set of Atari tasks from AGI-Labs\footnote{\url{https://github.com/AGI-Labs/continual_rl/blob/develop/docs/ATARI_RESULTS.md}}. In particular, the algorithms we consider include Continual Learning with Experience And Replay (CLEAR,~\cite{Rolnick2018clear}), online EWC, Progress \& Compress (P\&C,~\cite{Schwarz2018oewc}), and Importance Weighted Actor-Learner Architecture (IMPALA,~\cite{Espeholt2018impala}).

\begin{figure*}[t]
\centering
\includegraphics[width=.7\linewidth]{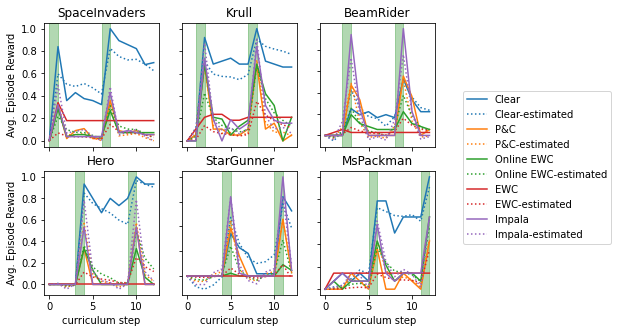}\caption{Performance of several common lifelong reinforcement learning approaches and baselines applied to a curriculum of  6 Atari tasks.  Each phase of the curriculum was trained for 5 million steps. Green vertical bars indicate the regions where the current task is being trained.  Dashed lines indicate the predicted performance curves estimates by \ourapproach. } \label{fig:atari_data}
\end{figure*}

Before running \ourapproach, we downsampled the performance data for each continual learning algorithm to capture task performance at task transition boundaries. We then normalized reward results across tasks using the procedure detailed in~\cite{New2022metrics}. The resulting data is shown in Figure \ref{fig:atari_data}.  We fit a \ourapproach model to the data as described in Section \ref{sec:estimation} resulting in the estimated algorithm performance data shown in \ref{fig:atari_data} ($0.007$ MSE).

The property estimates from \ourapproach shown in Tables~\ref{tab:rl_estimated_properties},~\ref{tab:rl_estimated_relationship}, and~\ref{tab:cifar_estimated_difficulties} qualitatively match well with our observations about the data shown in Figure \ref{fig:atari_data}. Of the approaches considered in this experiment, CLEAR was estimated to have the best transfer efficiency $0.12$ and experience retention $0.9$ due to the experience replay.  The data in Figure \ref{fig:atari_data} show that Clear frequently had the highest task performance and the highest task performance for tasks not actively being trained.  We observed that EWC and online EWC had relatively lower task performance, but they retained task performance while other tasks were training.  The CLAMP analysis explains the relatively poor performance of EWC and online EWC as being a result of relatively low transfer efficiency.  This makes sense as the mechanism of EWC and online EWC are based on regularization which adds additional loss terms that are simultaneously being optimized to slow parameter changes to parts of the network identified as useful for other tasks.  The algorithm parameter estimates from from CLAMP qualitatively match our expectations from the given performance data and a mechanistic understanding of the lifelong learning approaches.

\begin{table}[]
    \centering
    \subfloat[Estimated latent properties of different algorithms for based on Atari tasks]{
    \resizebox{.4\columnwidth}{!}{%
    \begin{tabular}{c||c|c|c}
    Approach                &   $\gamma$    &   $h$     &   $\lambda$ \\\hline\hline
    Clear                   &   0.12        &   0.90    &   0.03\\\hline
    Progress \& Compress    &   0.06        &   0.35    &   1.16\\\hline
    EWC\_online             &   0.03        &   0.60    &   0.82\\\hline
    EWC                     &   0.01        &   0.74    &   0.72\\\hline
    Impala                  &   0.10        &   0.33    &   1.26
    \end{tabular}
    }
    \label{tab:rl_estimated_properties}
    }
    \subfloat[Estimated task transfer between Atari environments]{
    \resizebox{.4\columnwidth}{!}{%
    \begin{tabular}{c||c|c|c|c|c|c}
    Task            &   \begin{tabular}{@{}c@{}}Space \\Invaders\end{tabular} &   Krull   &   \begin{tabular}{@{}c@{}}Beam \\Rider\end{tabular}  &   Hero    &   \begin{tabular}{@{}c@{}}Star\\ Gunner\end{tabular}     &   \begin{tabular}{@{}c@{}}Ms.\\ Packman\end{tabular}\\\hline\hline
    \begin{tabular}{@{}c@{}}Space\\ Invaders\end{tabular}  &   1.00            &   0.15    &   -0.13       &   0.01    &   -0.16           &   0.03\\\hline
    Krull           &   -0.09           &   1.00    &   0.21        &   -0.02   &   -0.05           &   0.02\\\hline
    \begin{tabular}{@{}c@{}}Beam \\Rider\end{tabular}      &   0.04            &   -0.15   &   1.00        &   0.01    &   0.04            &   0.08\\\hline
    Hero            &   0.13            &   0.05    &   -0.33       &   1.00    &   0.10            &   0.11\\\hline
    \begin{tabular}{@{}c@{}}Star\\ Gunner\end{tabular}     &   0.01            &   0.07    &   -0.16       &   -0.12   &   1.00            &   -0.03\\\hline
    \begin{tabular}{@{}c@{}}Ms.\\ Packman\end{tabular}     &   -0.03           &   -0.16   &   -0.12       &   -0.01   &   -0.38           &   1.00
    \end{tabular}
    }
    \label{tab:rl_estimated_relationship}
    }\\
    \subfloat[Estimated task difficulty for Atari tasks]{
    \resizebox{.4\columnwidth}{!}{%
    \begin{tabular}{c||c|c|c|c|c|c}
    Task            &   \begin{tabular}{@{}c@{}}Space \\Invaders\end{tabular} &   Krull   &   \begin{tabular}{@{}c@{}}Beam \\Rider\end{tabular}  &   Hero    &   \begin{tabular}{@{}c@{}}Star\\ Gunner\end{tabular}     &   \begin{tabular}{@{}c@{}}Ms.\\ Pacman\end{tabular}\\\hline\hline
    Difficulty  &   0.09    &   0.08    &   0.15    &   0.07    &   0.10    &   0.08\\
    \end{tabular}
    }
    \label{tab:rl_estimated_difficulty}
    }
    \caption{CLAMP estimates for the curriculum of Atari tasks show in Figure \ref{fig:atari_data} including (a) latent properties for lifelong learning algorithms including transfer efficiency $\gamma$, experience retention $h$, and expertise translation $\lambda$, (b) task transfer matrix indicating how experience on one task effects the others, and (c) intrinsic task difficulty.   }
\end{table}

\section{Discussion}

The analysis of synthetic data in Appendix \ref{sec:synthetic} demonstrates that the optimization approach in \ourapproach was able to recover underlying parameters of the model from synthetic lifelong learning performance data with low error.  From Section \ref{sec:atari} and Appendix \ref{sec:mnist} and \ref{sec:cifar100}, we found that \ourapproach had low mean squared error to real performance data from several benchmarks.  Qualitative assessment of the latent property estimates from analysis of the Atari experiment indicated broad agreement with expectations given the performance data and the underlying mechanisms of the lifelong learning approaches.

Existing metrics for evaluating lifelong learning conflate task structure with algorithm performance.  We have introduced \ourapproach - Continual Learning Analysis via a Model of Performance - as the first attempt to estimate the separate contributions of task structure from lifelong learning algorithm performance. \ourapproach  is model and task agnostic, and it can estimate several important properties of lifelong learning systems, including how well they learn from new experiences and how well they retain prior experience. We have also demonstrated the applicability of this approach to continual learning in both the classification and reinforcement learning settings.





\bibliography{references}
\bibliographystyle{collas2022_conference}

\appendix

\section{Related Work}
\label{sec:related}


\subsection{Explainable surrogate models}
\label{sec:explainable}
Explainable surrogate models are a common approach in the explainable AI (XAI, ~\cite{Angelov2021xai,Adadi2018model_agnostic,Arrieta2020xai2,darpaxai}) literature. These are a form of explanation by simplification~(\cite{Tritscher2020simplification}). By fitting a simplified model to a more complex model, by comparison, it may be possible to better interpret salient factors for prediction. For example, a linear model or decision tree might be fit to the original model to reduce the model complexity compared to the original model. \cite{doran2017xai} refer to this category of models as interpretable in that it is possible to mathematically analyze the algorithmic properties.

Another characteristic of explainable models are the assumptions that they make about knowledge of the original model structure.  Some approaches assume knowledge of the original model composition (white box assumption) while other approaches only assume that access to inputs and output are available from the full model (black box assumption)~(\cite{Adadi2018model_agnostic}).  Approaches that make black-box assumptions are also referred to as model-agnostic, in the sense that they can be used without detailed knowledge of the original models structure.  For example, Local Interpretable Model-agnostic Explanations~(\cite{Ribeiro2016Lime}) (LIME) is a recent example of a model agnostic approach that creates locally-optimized explanations by training a surrogate model.

Furthermore, there is a rich history of latent parameter inference in the Bayesian~(\cite{bayesian,vonToussaint2011bayesianphysics}) and generative model~(\cite{Zou2012generativelatent}) literature. Bayesian belief networks~(\cite{Holzinger2018bayesianxai}) have been explored for use in XAI.  

\subsection{Algorithms to mitigate catastrophic forgetting}
\label{sec:lifelong}
While lifelong learning attempts to address various shortcomings of modern AI, numerous lifelong learning algorithms pay special attention to \emph{catastrophic forgetting}~\cite{French1999CF,McCloskey1989CF}). Catastrophic forgetting is the behavior of an algorithm or model where, after training the model on one task, training on a new task adversely effects model's performance on the first task; often to the point that the model performs worse than a model behaving randomly. Nuance between how to calculate relevance and minimization of adverse effects differentiates algorithms such as \textit{elastic weight consolidation} (EWC,~\cite{Kirkpatrick2017ewc}), \textit{continual learning through synaptic intelligence} (SI,~\cite{Zenke2017si}), and \textit{memory aware synapses} (MAS,~\cite{Aljundi2018mas}). SI, for example, holds weights relatively fixed according to the sensitivity of loss function with respect to them, whereas MAS considers sensitivity of weights in the predictor itself. EWC, on the other hand, uses Fisher information for determining how to update weights. Gradient episodic memory (GEM,~\cite{LopezPaz2017gem}) like SI stores a subset of previous task data and seeks to minimize cost function on current labeled data point while constraining that loss on the episodic memory does not increase. 

\subsection{Continual learning experimental categories}\label{sec:categories}

We make use of two different continual learning experimental categories in this study which we briefly describe below.  They are referred to as domain incremental learning and task incremental learning.

\emph{Domain incremental learning} as defined by~\cite{Hsu2018strong_baselines} and~\cite{VanDeVen2018rtf} implies that task identifiers are not given to the lifelong learning algorithm.  The algorithm must infer from the given data which task is being performed.

\emph{Task incremental learning} is defined by~\cite{Hsu2018strong_baselines} as prediction where the task label is given to the algorithm.  Lifelong learning approaches for task adaptation often have an output prediction head for each task.  During inference, only the predictions from the current task head are considered. 

\section{Analysis of CLAMP with Synthetic Algorithm Performance Data}
\label{sec:synthetic}

We aimed to verify that our surrogate model functional form yields problems that are identifiable -- i.e., that a performance curve can be mapped back to a unique set of parameters. We designed an experiment using synthetic data where we could be certain of the underlying parameters governing the model.  We sampled ground truth parameters ($\mathbf{A}, d, \gamma, h,$ and $\lambda$) uniformly from $[-1, 1]$ for $\mathbf{A}$ and $[0, 1]$ for the other parameters, and we generated a random curriculum of length $9$ from a set of five tasks.  Then we used ${E}^a_j(c_l)$ and $\mathcal{P}_j(c_l)$ to generate synthetic lifelong learning performance data for 3 algorithms over a curriculum composed of $5$ tasks.  


\begin{figure*}[h!]
\centering
\includegraphics[width=0.65\linewidth]{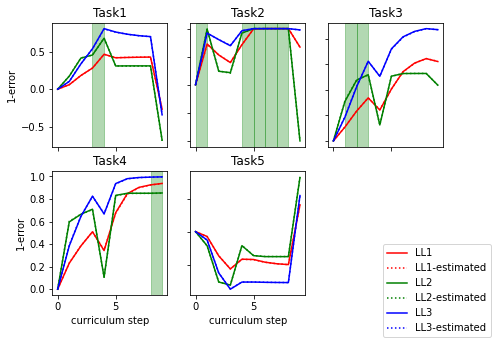}
\caption{Synthetic performance data and the fit of the surrogate model to the data. Green bars on the $x$-axis for a given task indicate periods in the curriculum where that task was trained. Our curriculum was randomly generated, so Task 5 was never trained on. In all cases, the estimated performance curves (dashed lines) overlap very closely to the synthetic performance data (solid lines).}
\label{fig:synthetic_data}
\end{figure*}

\begin{table}[h!]
    \centering
    \caption{Parameter distributions and estimation errors for synthetic scenario generation experiment}
    \resizebox{.4\columnwidth}{!}{%
    \begin{tabular}{c||c|c|c}
                                       &    Symbol         & Ranges & MSE\\\hline\hline
        \textbf{Task Parameters}       &                   &                 & \\\hline
        Transfer Matrix                & $\mathbf{A}$      & $[-1, 1]$       & 0.12\\\hline
        Difficulty                     & $d$               & $[0,\infty)$    & 0.04\\\hline
                                       &                   &                 & \\\hline
        \textbf{Algorithm Parameters}  &                   &                 & \\\hline
        Transfer Efficiency            & $\gamma$          & $[0,\infty)$    & 0.02\\\hline
        Experience Retention           & $h$               & $[0, 1]$        & 0.0\\\hline
        Expertise Translation          & $\lambda$         & $[0,\infty)$    & 0.01
    \end{tabular}
    }
    \label{tab:synthetic}
\end{table}


To validate the model fitting approach to recover underlying model parameters, we fit a randomly initialized surrogate performance model to the synthetic performance data for 1000 epochs using an Adam optimizer~(\cite{Kingma2014adam}) with mean squared error as the loss function. The estimated curves match the synthetic data with low error as expected.  
We then computed the mean squared error between the ground truth parameters and the parameters in the fitted model.  The results are shown in Table~\ref{tab:synthetic}. We can conclude that the approach used to fit the explainable performance surrogate model can be used to recover accurate estimates of the true underlying parameter values with small error.



\section{Analysis of MNIST and CIFAR100 Lifelong Learning Performance}
\label{sec:classification}

Here we evaluate whether the strong assumptions we used to develop our model of learning can describe real lifelong learning performance data for classification algorithms. We designed experiments including the MNIST and CIFAR100 datasets to quantitatively evaluate how well our model fits real lifelong performance data. To generate performance data, we followed the experimental protocol described by~\cite{Hsu2018strong_baselines}.

Performance curves from four classification algorithms were considered. The first (NormalNN) makes use of a neural network where weights are transferred from one task to the next.  A second (L2) included $\ell^2$ regularization in its loss function to minimize the change in weights from those that were previously learned. Naive rehearsal (Naive\_Rehearsal\_[k]) was a third algorithm that made use of an experience replay memory to store and retrain on a subset of size $k$ of previously experienced data at random. Each training batch was composed of equal parts from the current task and the experience replay memory. The number of stored experiences over all tasks is a parameter of the approach. 


We also included curves from four state-of-the-art lifelong learning algorithms: elastic weight consolidation (EWC,~\cite{Kirkpatrick2017ewc}), online elastic weight consolidation (EWC\_online,~\cite{Schwarz2018oewc}), synaptic intelligence (SI,~\cite{Zenke2017si}), memory-aware synapses (MAS,~\cite{Aljundi2018mas}), and gradient episodic memory (GEM,~\cite{LopezPaz2017gem}).



\subsection{Domain adaptation with Split-MNIST}
\label{sec:mnist}
For MNIST, we considered the problem of domain-incremental learning. We followed the experimental protocol described by~\cite{Hsu2018strong_baselines}, which we briefly describe below.

\begin{figure*}[h!]
\includegraphics[width=1\linewidth]{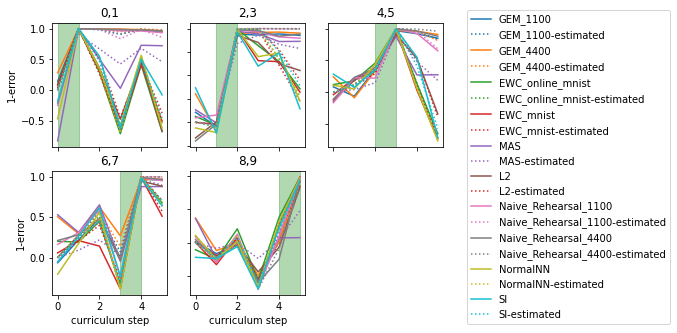}
\caption{Continual learning performance for incremental domain learning on Split-MNIST along with performance curves estimated with \ourapproach\hspace{-4pt}.  Naive rehearsal was considered with experience storage limits of $1100$ and $4400$.
}\label{fig:mnist_data}
\end{figure*}

We used the dataset splitting procedure to create multiple binary classification tasks (e.g., distinguish between images of 0 and images of 1) from the MNIST dataset with 60k training images ($\sim$6,000 per digit) and 10,000 test images. The 32x32 greyscale images were normalized to have a mean of zero and variance of one.

\begin{table}[]
    \centering
    \caption{Estimated latent properties of different algorithms for the MNIST task-incremental learning experiment}
    \resizebox{.4\columnwidth}{!}{%
    \begin{tabular}{c||c|c|c}
    Approach                &   $\gamma$    &   $h$     &   $\lambda$ \\\hline\hline
    GEM\_1100               &   0.79        &   0.99    &   0.30\\\hline
    GEM\_4400               &   0.88        &   1.00    &   0.11\\\hline
    EWC\_online             &   0.96        &   0.44    &   0.58\\\hline
    EWC                     &   0.86        &   0.44    &   0.56\\\hline
    MAS                     &   0.32        &   0.91    &   0.09\\\hline
    L2                      &   0.74        &   0.53    &   0.65\\\hline
    Naive\_Rehearsal\_1100  &   1.02        &   0.89    &   0.35\\\hline
    Naive\_Rehearsal\_4400  &   0.92        &   0.97    &   0.63\\\hline
    NormalNN                &   0.99        &   0.45    &   0.53
    \end{tabular}
    }
    \label{tab:mnist_estimated_properties}
\end{table}

\begin{table}[]
    \centering
    \caption{Estimated task transfer scores for MNIST task-incremental learning experiment}
    \resizebox{.4\columnwidth}{!}{%
    \begin{tabular}{c||c|c|c|c|c}
    Task    &   0,1     &   2,3     &   4,5     &   6,7     &   8,9\\\hline\hline
    0,1     &   1.0     &   0.0     &   0.03    &   0.05    &   -0.1\\\hline
    2,3     &   -0.4    &   0.88    &   0.07    &   0.07    &   0.12\\\hline
    4,5     &   -0.2    &   0.38    &   0.99    &   -0.1    &   -0.4\\\hline
    6,7     &   0.27    &   -0.2    &   -0.7    &   1.0     &   0.4\\\hline
    8,9     &   -0.2    &   0.0     &   -0.2    &   -0.2    &   0.75
    \end{tabular}
    }
    \label{tab:mnist_estimated_relationships}
\end{table}


The results of our MNIST analysis are given in Fig.~\ref{fig:mnist_data} (estimated and true performance curves) and Tables~\ref{tab:mnist_estimated_properties} (estimated properties) and~\ref{tab:mnist_estimated_relationships} (estimated transfer scores). Averaged over each classification algorithm $a$, the surrogate models achieved a squared error $\text{mean}_a||\hat{P}^a - P^a||_F^2$ of $0.005$. The ability of the surrogate model to fit the performance data with low error is a partial validation of the formulation of the surrogate performance model. Table~\ref{tab:mnist_estimated_relationships} is the estimated task transfer matrix. The estimated task transfer diagonal was strongly positive.  The highest off-diagonal transfer estimate was that training with the $6,7$ task is beneficial for the task $8,9$ task.  We believe this result follows intuition because of the character similarity between $6$ and $9$.





The estimated algorithm properties are shown in Table~\ref{tab:mnist_estimated_properties}.  At first glance it might seem surprising that a normal neural network (Normal NN) had one of the highest estimated transfer efficiencies $\gamma$ for the MNIST set of tasks, given that we know that normal neural networks are ill-suited for continual learning.  One intuition for the result is that normal neural networks are able to transfer well because their weight updates are unconstrained, while the regularization in other methods prevents such rapid transfer. However, because the normal neural network updates are unconstrained, performance on past tasks is not maintained, while the regularized methods are able to maintain performance on various tasks throughout the curricula.


\subsection{CIFAR100 task-incremental learning} 
\label{sec:cifar100}
For analysis on CIFAR100, we used results from \emph{task-incremental learning}.  We used the protocol by~\cite{Hsu2018strong_baselines} for preparing the CIFAR100 dataset and running the continual learning approaches.  

Like the MNIST-task set, we used the dataset splitting procedure to create 5 binary classification tasks based on the CIFAR100 dataset.  The 100 CIFAR classes were subdivided at random into 10 classes.  Each binary classification task consisted of two of the 10 distinct classes.  Data preprocessing resulted in 32x32 images that were normalized with a mean of zero and variance of one.  To ensure a fair comparison among lifelong learning approaches, all lifelong learning approaches used the same WideResNet~(\cite{Zagoruyko2016wideresnet}) architecture.

\begin{figure*}[h!]
\includegraphics[width=\linewidth]{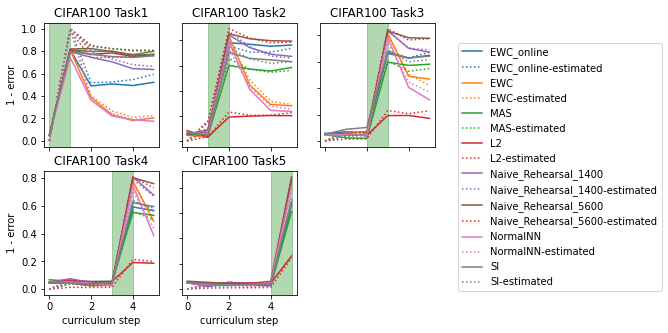}
\caption{Continual learning performance for incremental task learning on Split-CIFAR100 tasks of several lifelong learning algorithms along with the performance estimated with \ourapproach\hspace{-4pt}.  Naive rehearsal was considered with experience storage limits of $1400$ and $5600$.}\label{fig:cifar_data}
\end{figure*}

The solid lines in Figure \ref{fig:cifar_data} show the performance (i.e. 1-error) for each of the tasks across the curriculum.  The highlighted green regions in the figure illustrate when the task is actively being trained.  As expected, we broadly observe that performance of a task increases sharply when actively being trained.

The fit of the surrogate performance model is shown as dashed lines in Figure \ref{fig:cifar_data}, and illustrates broad general agreement between predicted performance and the underlying data.  We observed a MSE of $0.01$ between the predicted and observed performance curves.  


\begin{table}[]
    \centering
    \caption{Estimated latent properties of different algorithms for the CIFAR100 task-incremental learning experiment}
    \resizebox{.4\columnwidth}{!}{%
    \begin{tabular}{c||c|c|c}
    Approach                &   $\gamma$    &   $h$     &   $\lambda$  \\\hline\hline
    EWC\_online             &   0.35        &   1.0     &   2.11\\\hline
    EWC                     &   0.60        &   0.61    &   0.00\\\hline
    MAS                     &   0.33        &   0.94    &   1.04\\\hline
    L2                      &   0.12        &   1.00    &   0.00\\\hline
    Naive\_Rehearsal\_1400  &   0.57        &   0.84    &   1.03\\\hline
    Naive\_Rehearsal\_5600  &   0.58        &   0.91    &   1.37\\\hline
    NormalNN                &   0.52        &   0.62    &   0.00\\\hline
    SI                      &   0.39        &   0.93    &   1.21
    \end{tabular}
    }
    \label{tab:cifar_estimated_properties}
\end{table}

\begin{table}[]
    \centering
    \caption{Estimated task transfer scores for CIFAR100 task-incremental learning experiment}
    \resizebox{.4\columnwidth}{!}{%
    \begin{tabular}{c||c|c|c|c|c}
    CIFAR100 Task   &   1       &   2       &   3       &   4       &   5\\\hline\hline
    1               &   0.96    &   0.15    &   0.06    &   0.07    &   0.04\\\hline
    2               &   -0.56   &   0.91    &   0.01    &   -0.01   &   0.01\\\hline
    3               &   0.00    &   -0.13   &   1.00    &   0.01    &   0.00\\\hline
    4               &   0.00    &   0.00    &   -0.12   &   1.00    &   0.01\\\hline
    5               &   0.01    &   0.09    &   0.10    &   -0.09   &   1.00
    \end{tabular}
    }
    \label{tab:cifar_estimated_relationships}
\end{table}

\begin{table}[]
    \centering
    \caption{Estimated task difficulty values for CIFAR100 task-incremental learning experiment}
    \resizebox{.4\columnwidth}{!}{%
    \begin{tabular}{c||c|c|c|c|c}
    CIFAR100 Task   &   1       &   2       &   3       &   4       &   5\\\hline
    Difficulty      &   0.0     &   0.3     &   0.3     &   0.3     &   0.3\\
    \end{tabular}
    }
    \label{tab:cifar_estimated_difficulties}
\end{table}


In Table~\ref{tab:comparison}, we compared the estimated transfer efficiency of multiple algorithms between the MNIST domain and CIFAR-100 domain.  Each column was colored from green to red relative to the values in the column.  Between the MNIST and CIFAR100 experiment there is broad agreement about the rank ordering of algorithms with respect to transfer efficiency. This indicates that latent property analysis might be useful to estimate the relative performance of a lifelong learning algorithm from one dataset to another.



\begin{table}[]
    \centering
    \caption{Estimated transfer efficiencies scores for different algorithms across MNIST and CIFAR100 task-incremental learning experiments}
    \resizebox{.4\columnwidth}{!}{%
    \begin{tabular}{c||c|c}
    Algorithm               &   MNIST   &   CIFAR100\\\hline\hline
    EWC\_online             &   0.96    &   0.35\\\hline
    EWC                     &   0.86    &   0.60\\\hline
    MAS                     &   0.32    &   0.33\\\hline
    L2                      &   0.74    &   0.12\\\hline
    Naive\_Rehearsal\_Low   &   1.02    &   0.57\\\hline
    Naive\_Rehearsal\_High  &   0.92    &   0.58\\\hline
    NormalNN                &   0.99    &   0.52\\\hline
    SI                      &   0.85    &   0.39
    \end{tabular}
    }
    \label{tab:comparison}
\end{table}


\subsection{Additional Discussion} 
\label{sec:discussion}

Our contributions include: (i) the introduction of the first explainable algorithm-agnostic surrogate performance model of lifelong learning, (ii) quantitative validation of the optimization procedure used to estimate the metrics, (iii) qualitative validation of the lifelong learning surrogate performance model using data produced by several popular lifelong learning approaches on benchmark datasets including CIFAR100, MNIST, and Atari.

Our approach to task transfer score recovery is dataset and algorithm-agnostic, and therefore generally applicable to all domains of lifelong learning. The algorithm latent properties seem to be predictive of how the algorithm may perform on other datasets.

A limitation of \ourapproach\,\,is that it makes a number of assumption about lifelong learning with the goal of explaining in part the performance of lifelong learning algorithms over diverse tasks.  For example, we assume linear experience transfer between tasks.  In future work, it would be interesting to explore relaxing the linear task transfer assumptions and consider non-linear transfer.

\end{document}